\documentclass[preprint,12pt]{elsarticle}

\usepackage{amssymb}
\usepackage{graphicx}
\usepackage{caption}
\usepackage{subcaption}
\usepackage{xcolor}
\usepackage{tabularray}
\usepackage{todonotes}
\usepackage{amsmath}
\usepackage{float}
\usepackage{makecell}
\usepackage{url}
\UseTblrLibrary{booktabs}
\usepackage{comment}
\usepackage{pifont}
\newcommand{\cmark}{\ding{51}}%
\newcommand{\xmark}{\ding{55}}%

\urldef{\mailsa}\path|{grios, pdalbianco, fronchetti, fquiroga, whasperue}@lidi.info.unlp.edu.ar|

\journal{Applied Soft Computing}

\begin{document}

\begin{frontmatter}



\title{HandCraft: Dynamic Sign Generation for Synthetic Data Augmentation}


\author[lidi,infounlp]{Gaston Gustavo Rios\corref{cor1}}
\ead{grios@lidi.info.unlp.edu.ar}
\author[lidi,infounlp]{Pedro Dal Bianco}
\author[lidi,cic]{Franco Ronchetti}
\author[lidi,cic]{Facundo Quiroga}
\author[lidi,conicet]{Oscar Stanchi}
\author[lidi,infounlp]{Santiago Ponte Ahón}
\author[lidi]{Waldo Hasperué}

\affiliation[lidi]{organization={Instituto de Investigación en Informática LIDI - Universidad Nacional de La Plata},
            addressline={50 \& 120}, 
            city={La Plata},
            postcode={1900}, 
            state={Buenos Aires},
            country={Argentina}}

\affiliation[cic]{organization={Comisión de Investigaciones Científicas de la Provincia de Buenos Aires (CICPBA)},
            city={La Plata},
            postcode={1900}, 
            state={Buenos Aires},
            country={Argentina}}

\affiliation[infounlp]{organization={Becario Doctoral - Universidad Nacional de La Plata},
            addressline={50 \& 120}, 
            city={La Plata},
            postcode={1900}, 
            state={Buenos Aires},
            country={Argentina}}

\affiliation[conicet]{organization={Becario Doctoral, CONICET},
            city={La Plata},
            postcode={1900}, 
            state={Buenos Aires},
            country={Argentina}}

\cortext[cor1]{Corresponding author.}

\begin{abstract}


Sign Language Recognition (SLR) models face significant performance limitations due to insufficient training data availability. In this article, we address the challenge of limited data in SLR by introducing a novel and lightweight sign generation model based on CMLPe. This model, coupled with a synthetic data pretraining approach, consistently improves recognition accuracy, establishing new state-of-the-art results for the LSFB and DiSPLaY datasets using our Mamba-SL and Transformer-SL classifiers. 
Our findings reveal that synthetic data pretraining outperforms traditional augmentation methods in some cases and yields complementary benefits when implemented alongside them. Our approach democratizes sign generation and synthetic data pretraining for SLR by providing computationally efficient methods that achieve significant performance improvements across diverse datasets.

\end{abstract}

\begin{graphicalabstract}
\includegraphics[width=\hsize]{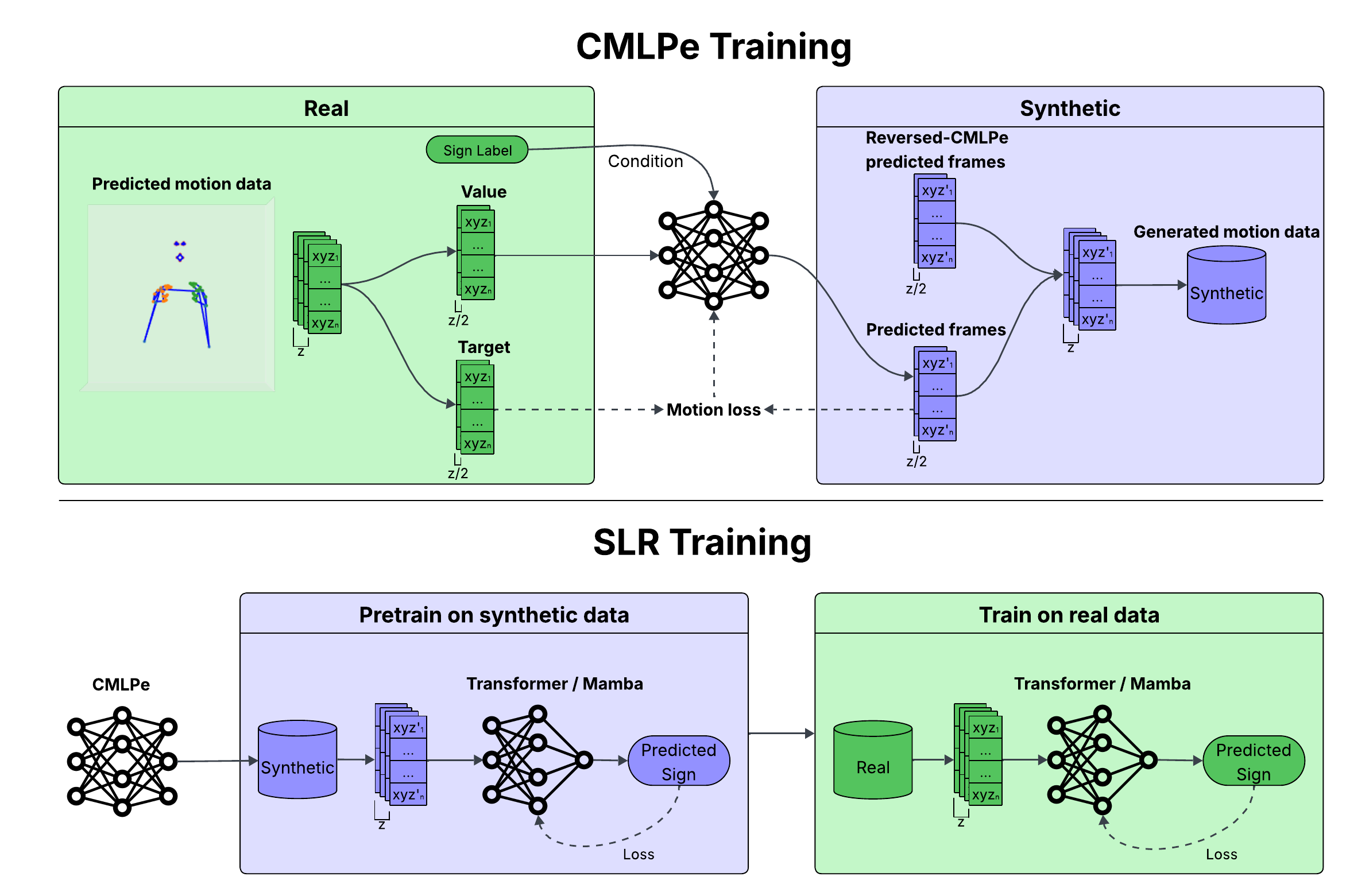}
\end{graphicalabstract}

\begin{highlights}
\item Created CMLPe, a lightweight conditional sign pose generator
\item Synthetic data pretraining consistently improves SLR accuracy across datasets
\item State-of-the-art results for LSFB (55.0\%) and DiSPLaY (97.3\%) datasets
\item Complementary benefits when combining synthetic data with traditional augmentation
\end{highlights}

\begin{keyword}
Sign Language Generation \sep Human Motion Generation \sep Sign Language Recognition \sep Limited Data \sep Synthetic Data Augmentation


\end{keyword}

\end{frontmatter}





\begin{figure}[ht!]
    \centering
    \includegraphics[width=\hsize]{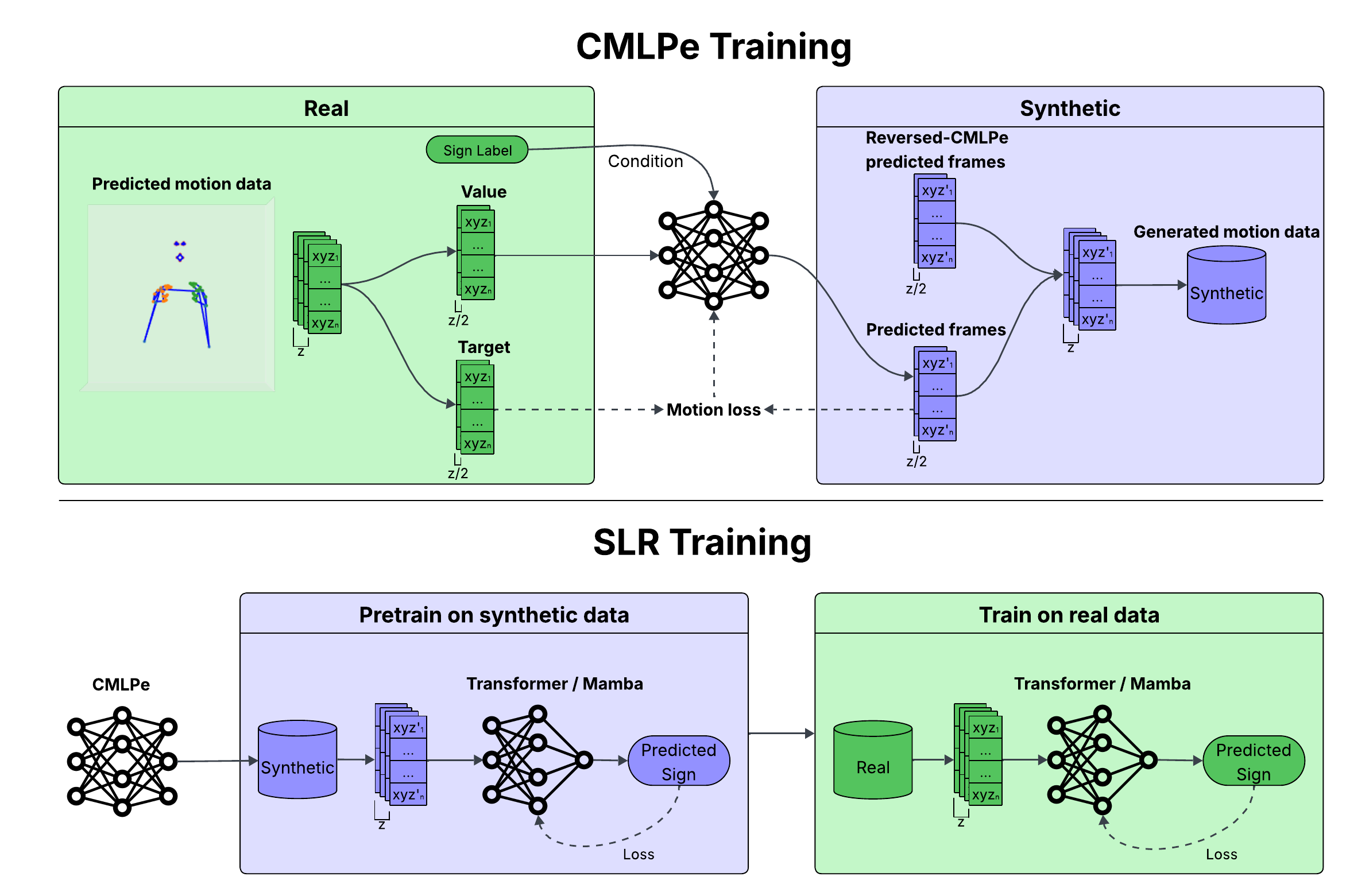}
    \caption{
    Our two-phase approach to sign language recognition using generated data. First, the CMLPe Training phase creates synthetic motion data by predicting different segments of sign movements using value/target pairs from real data. Second, the sign language recognition Training phase leverages this synthetic data to pretrain Transformer-S/Mamba-S networks before fine-tuning on real sign data, enhancing recognition performance through pretraining with generated data.}
    \label{fig:graphicalabstract}
\end{figure}

\section{Introduction}

Approximately 5\% of the global population experiences disabling hearing loss. By 2050 this number is estimated to grow to 10\% of the population  – or 700 million people –\cite{Hapunda2024}. Deafness is defined as a profound hearing loss that precludes an individual from processing linguistic information through hearing \cite{Elzouki2012}. Thus, deaf people often communicate via sign language. This visual language is expressed through face, body, and hand movements. Machine learning approaches offer promising solutions for developing automated translation systems between spoken and sign languages.

In recent years, Large Language Models (LLMs) and speech recognition models have achieved significant success in recognizing and translating numerous written or spoken languages. However, these models require extensive training data and their success does not readily transfer to Sign Language Recognition (SLR) due to fundamental differences in data availability, modality, and structural complexity. Furthermore, there exist more than 300 mutually intelligible sign languages worldwide \cite{NationalGeographic2025} and nearly 80\% of people with disabling hearing loss live in low and middle-income countries. This makes data availability a common problem for multiple sign languages.

There are various approaches that can help deal with low data availability. Data augmentation is the process of creating new training data in order to increase a deep learning model's generalization and avoid overfitting. Typically, this process is applied by introducing slight modifications to existing real training data without affecting the label. However, in scenarios where the data is limited and the model must perform inference under a varied set of conditions, with no practical way to access further data, synthetic data can supply this demand \cite{Mumuni2024}. 

Pretraining using generated data involves leveraging synthetically created samples to augment training datasets for deep learning models, effectively addressing limited or imbalanced data scenarios. This method has been proven effective in enhancing the training of models, resulting in faster and more stable convergence \cite{Mostofi2024}.

Sign Language Generation (SLG), translating from text to sign data, can provide an effective synthetic data source for SLR. SLG can be achieved by leveraging human motion synthesis to exploit its proven techniques, which are highly developed due to their application to graphics, gaming, and simulation environments for robotics. These architectures often use pose-based methods for their simplicity and effectiveness due to the lower dimensionality of poses when compared to video input. 

\subsection{Proposed approach}

Our approach to SLR aided by SLG is summarized in Figure \ref{fig:graphicalabstract}. We develop \textbf{Conditional MultiLayer Perceptron (CMLPe)} architectures that generate semantically consistent pose sequences conditioned on sign labels while preserving both temporal dynamics and spatial relationships. We based our generator model on the siMLPe \cite{Guo2022} multi-layer perceptron architecture. Each model generates signs for a different sign language: Indian Sign Language (INCLUDE), French Belgian Sign Language (LSFB) and Medical Sign Language (DiSPLaY). Each sign language dataset encompasses different characteristics, with a wide variety of samples per class and unique signs.

These SLG models enable systematic synthetic dataset creation through class-balanced sampling strategies. This data is then used to pretrain SLR models in order to improve their effectiveness in limited and unbalanced data environments. The models are then fine-tuned on the real data distribution, applying data augmentation (random rotations and scaling variations) to further combat the lack of data quality. 

To compare the ability of augmenting the training data with generated samples in SLR, we developed two models: an encoder-only Transformer \cite{Vaswani2023} model (\textbf{Transformer-SL}) and a Mamba \cite{mamba} based classifier (\textbf{Mamba-SL}). These state-of-the-art architectures extract the temporal and spatial patterns of the pose sequences in order to get the correct sign by using self-attention and state space model blocks respectively.

Our methodology is designed to address the data scarcity in sign language recognition while maintaining computational efficiency. By focusing on pose data rather than raw video, we significantly reduce input dimensionality while preserving the essential spatial and temporal information needed for accurate recognition. Our lightweight generative model enables practical deployment even in research environments with limited computing resources.

\subsection{Contributions}

This work makes several significant contributions to the fields of sign language generation and recognition:

\begin{itemize}
    \item We develop CMLPe, a new conditional motion generator specifically designed for sign language data that creates diverse, semantically consistent synthetic samples while requiring minimal computational resources.
    \item We develop Transformer-SL and a Mamba-SL, two SLR models with state-of-the-art performance that leverage on both the spatial and temporal dimensions of the keypoints.
    \item We demonstrate that pretraining using generated data consistently improves recognition performance across three diverse sign language datasets (INCLUDE, LSFB, and DiSPLaY), establishing a new state-of-the-art for the LSFB and DiSPLaY datasets.
    \item We verify that our approach provides benefits beyond traditional data augmentation while demonstrating additive effectiveness when combined with it.
\end{itemize}

The code and models used in this research are available at:

\url{https://github.com/okason97/HandCraft}

\section{Related Work}

\subsection{Sign Language Recognition}

State-of-the-art SLR and gesture recognition commonly employ models based on convolutional neural networks \cite{Shiwei2024}, transformer architectures \cite{Wong2024} and combinations of the two architectures \cite{Skobov2023}. Due to the data limitations of sign language datasets, innovative data representation methods and training pipelines have been developed to enhance these models. Pose information extracted by pose recognition models has shown great success in improving performance \cite{Skobov2023}. This can be attributed to a better representation of the input data, retaining sufficient discriminative information to classify the signs while removing task-irrelevant information. Discrete Cosine Transform (DCT) \cite{Guo2022} has been successfully implemented to improve the representation of the data, encoding temporal information into it.

\subsection{Data Augmentation and synthetic data}

Data augmentation has emerged as an essential technique for enhancing model robustness and mitigating overfitting \cite{Rebuffi2021}. Furthermore, data augmentation can diminish the representation distance between video and text data, easing data scarcity. However, traditional data augmentation approaches are inherently constrained, as they can only marginally expand the original data distribution through minor transformations while preserving class labels. 

Synthetic data generation offers a more powerful alternative by creating entirely new samples rather than merely transforming existing ones \cite{Mumuni2024}. Two primary methodologies exist for generation: deep learning-based approaches and algorithmic simulation. The choice between these methods depends critically on data complexity and simulation feasibility.

Algorithmic simulators provide reliable and consistent data generation but are limited by implementation complexity. Creating comprehensive simulators for intricate data distributions requires substantial engineering effort, constraining their applicability to relatively simple tasks or domains where insufficient data precludes training robust generative models \cite{Yang2023}. A notable example in sign language recognition demonstrates this approach's effectiveness: skeletal-rigged 3D hand avatar meshes programmed to generate synthetic variations at scale successfully addressed data scarcity issues \cite{fowley2021sign}. 

In contrast, deep learning-based synthetic data generation offers broader applicability and seamless integration into existing training pipelines. However, this approach typically produces lower-quality samples that require careful handling to ensure effective utilization in model training. Generative adversarial networks \cite{Chen2024} and Denoising Diffusion models \cite{Dhawan2024} represent the predominant architectures for this generation paradigm.

\subsection{Sign Language Deep Learning-based Generation}

Recent advancements in deep learning have significantly improved the ability to generate realistic human poses in various contexts, from virtual agents to video conferencing and media production \cite{Sha2023}. These techniques typically leverage auto-regressive deep neural networks to learn the complex relationships between human body structure and natural movement patterns \cite{Dang2024}. The abstraction of the poses provides a simpler task than full image generation by removing unnecessary environmental information related to the motion of the human body. 

Sign language generation presents unique challenges beyond general human pose modeling, requiring precise hand gestures, facial expressions, and temporal coherence to convey linguistic meaning effectively. Sign language generation models often struggle with fine details \cite{Ventura2021} and temporal anomalies \cite{Shi2024}. As with human motion generation, the abstraction of the pose can provide a significant advantage, further aiding in the generation process \cite{Shi2024} or avatar animation \cite{Shalev2022}. Current works in SLG have shown a big improvement, especially when generating from a strong prior. In particular, the most successful of these generators work by extracting semantic information from already existing real sign language videos and generating new samples conditioned on this data \cite{Suo2024}. These, however, still suffer from a lack of realism in some areas, especially in the hands and face features, when generating the full frames from scratch, with some of these opting for 3D avatars to circumvent this challenge \cite{Dong2024}. 

\section{Datasets}

\begin{figure}[ht!]
    \centering
    \begin{subfigure}{0.3\textwidth}
        \includegraphics[width=\hsize]{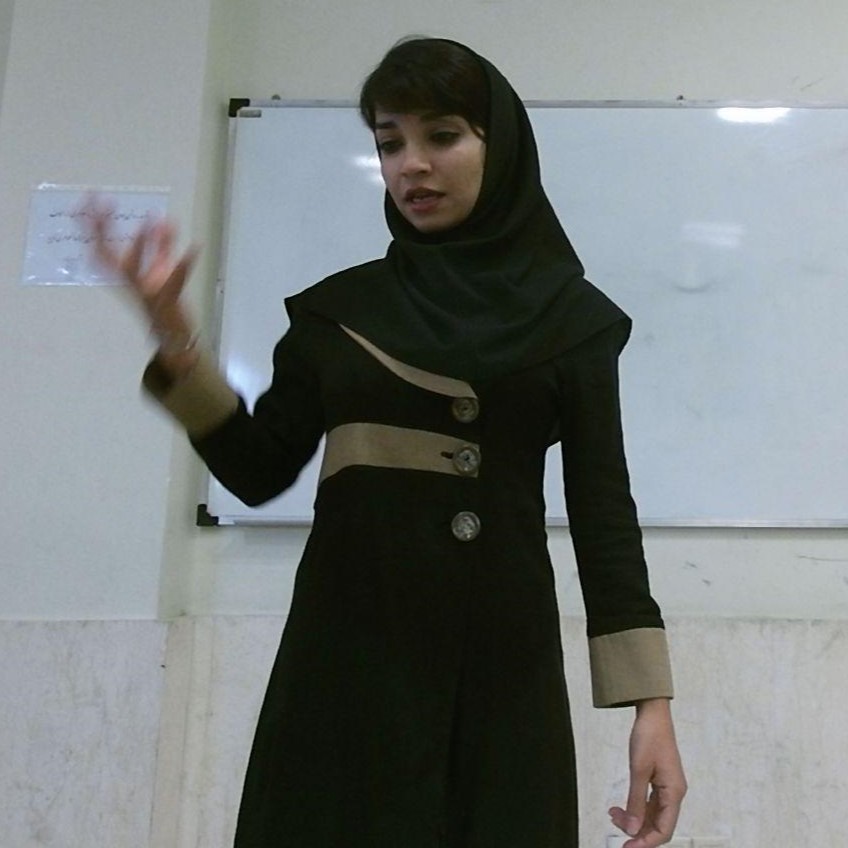}
        \caption{DiSPLaY}
        \label{fig:example_display}
    \end{subfigure}
    \hfill
    \begin{subfigure}{0.3\textwidth}
        \includegraphics[width=\hsize]{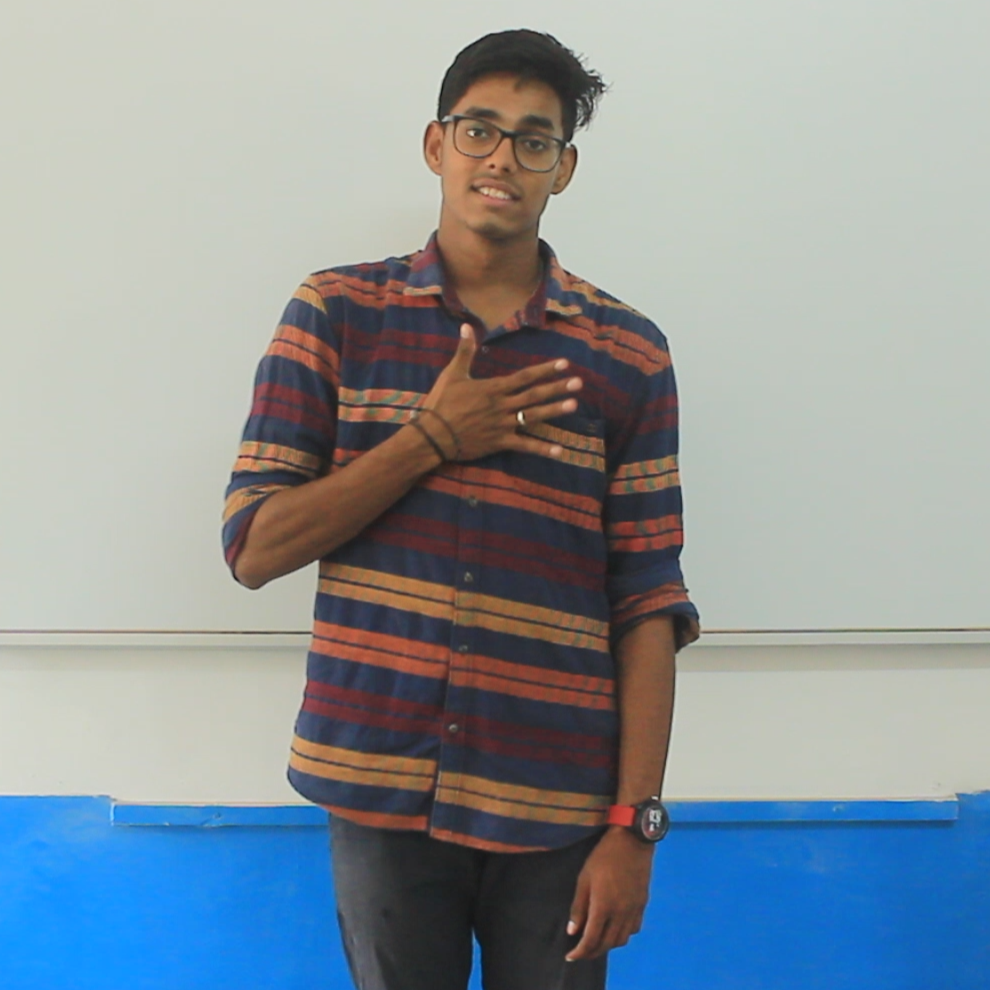}
        \caption{INCLUDE}
        \label{fig:example_include}
    \end{subfigure}
    \hfill
    \begin{subfigure}{0.3\textwidth}
        \includegraphics[width=\hsize]{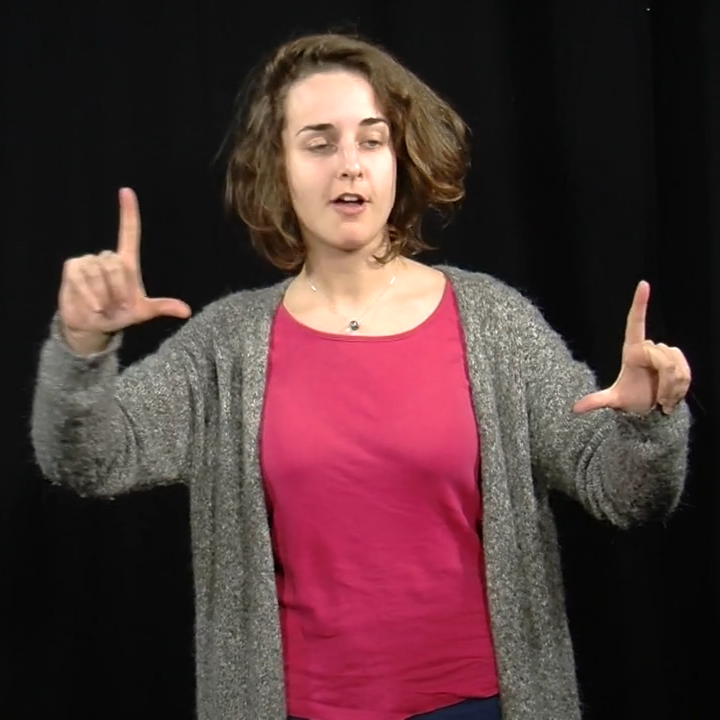}
        \caption{LSFB}
        \label{fig:example_lsfb}
    \end{subfigure}
    \begin{subfigure}{0.3\textwidth}
        \includegraphics[width=\hsize]{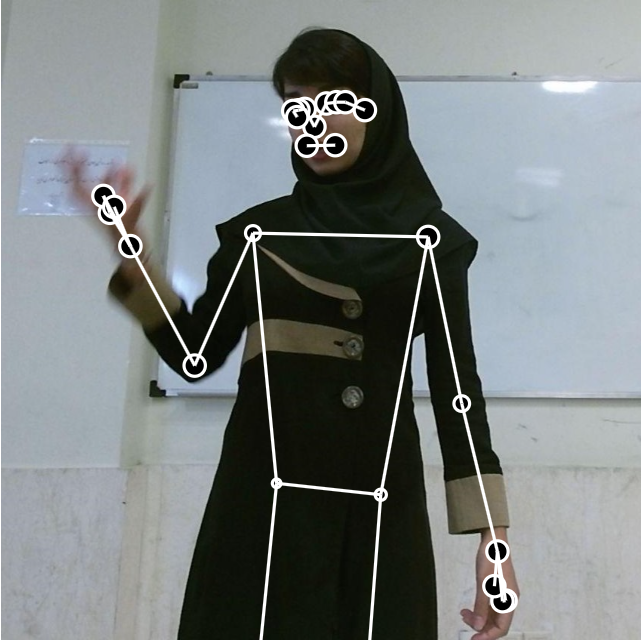}
        \caption{DiSPLaY keypoints}
        \label{fig:example_display_pose}
    \end{subfigure}
    \hfill
    \begin{subfigure}{0.3\textwidth}
        \includegraphics[width=\hsize]{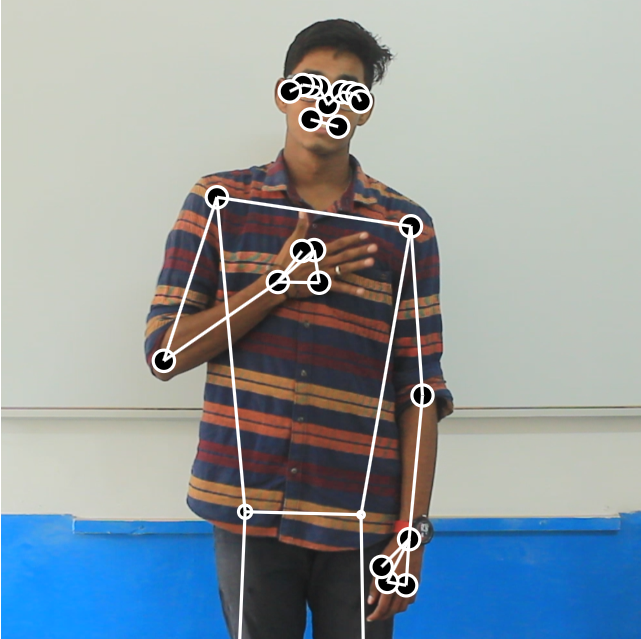}
        \caption{INCLUDE keypoints}
        \label{fig:example_include_pose}
    \end{subfigure}
    \hfill
    \begin{subfigure}{0.3\textwidth}
        \includegraphics[width=\hsize]{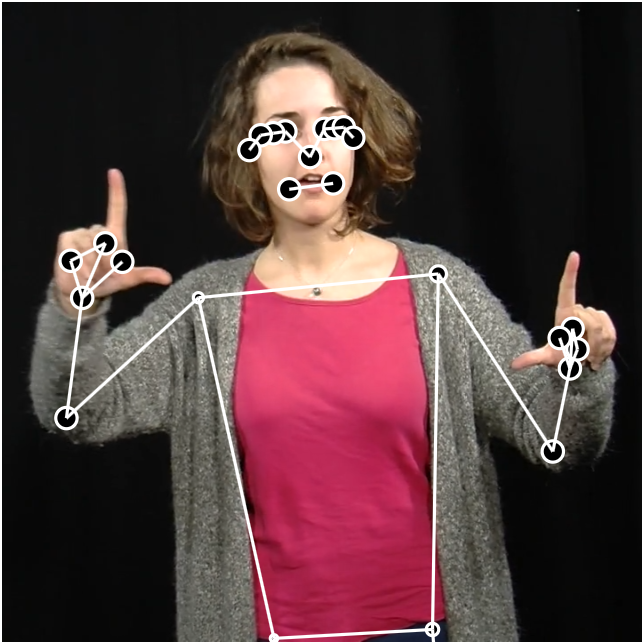}
        \caption{LSFB keypoints}
        \label{fig:example_lsfb_pose}
    \end{subfigure}
    \caption{Visualization of images from the Display (a, d), INCLUDE (b, e) and LSFB (c, f) datasets before and after keypoints extraction.}
    \label{fig:datasets_visualization}
\end{figure}

To validate our methodology, we conducted experiments on a diverse set of datasets with distinct characteristics. DiSPLaY, LSFB and INCLUDE encompasses different sign languages, enabling us to evaluate our approach across related but non-identical domains. The varying sizes of these datasets—ranging from the relatively small DiSPLaY to the more comprehensive LSFB corpus—allow us to assess our models' performance under different data availability conditions. Furthermore, the substantial variations in sequence lengths and class counts across all datasets introduce additional complexity to our experimental evaluation. The details of each dataset can be found in Table \ref{tab:datasets} and samples in Figure \ref{fig:datasets_visualization}. We use the testing samples provided by the dataset authors and allocate 10\% of the training samples for validation. To address class imbalance during generation and recognition, we implemented oversampling techniques, equalizing the sample count across all classes to match the most populated category.

In this paper, we focus solely on the poses, as they reduce domain complexity and enable faster processing times for models. We extracted pose data using the MediaPipe tool and subsequently normalized it. This resulted in 478 3D landmarks for the face, 33 for the body, and 21 2D landmarks for each hand. We selected a subset of 12 face landmarks of the eyes and mouth, and 6 body keypoints to reduce data dimensionality while preserving discriminative information. This totals 60 landmarks for our input, as can be seen in Figure \ref{fig:datasets_visualization}. We further processed the keypoints to avoid sequence discontinuity and mitigate noise between frames given by the lack of temporal precision of the model. Linear interpolation was used to fill the missing values. Savitzky-Golay filter was used with a window of 15 and a polynomial order of 3 in order to smooth values and mitigate noise.

\begin{table}
\centering
\setlength{\tabcolsep}{4pt} 
\footnotesize 
\caption{Comparison of sign language datasets used in our experiments. The table shows each dataset's sign language (Sign Language), number of unique signs (Unique signs), sample distribution between training and testing sets (Samples), sequence length range in frames (Seq. length), and the range of samples per sign class (Class count).}
\makebox[\linewidth]{
    \begin{tabular}{lcccccc}
        \hline
        \multicolumn{1}{c}{\textbf{Name}} & \textbf{Sign Language} & \textbf{Unique signs} & \textbf{Samples} & \textbf{Seq. length} & \textbf{Class count} \\ \hline
        \textbf{DiSPLaY} & Medical & 55 & \makecell{1,209 train + \\ 515 test} & 64-172 & 4-21 \\
        \textbf{INCLUDE} & Indian & 253 & \makecell{2,949 train + \\ 908 test} & 34-154 & 2-20 \\
        \textbf{LSFB} & French Belgian & 610 & \makecell{52,350 train + \\ 39,831 test} & 4-60 & 1-1,514 \\ \hline
    \end{tabular}
}
\label{tab:datasets}
\end{table}

\subsection{LSFB}

The French Belgian Sign Language Isolated (LSFB-ISOL) \cite{Fink2021} dataset is built upon the LSFB Corpus. It spans 25 hours of videos and poses of continuous isolated signs performed by 85 different signers. After filtering out signs with less than 20 samples or more than 60 frames, the dataset contained 52,350 sign poses across 610 classes.

The high number of classes and high class imbalance with low sequence length makes this a challenging dataset for SLR. The models must be able to work with varying sample lengths, including some samples that contain as few as 4 frames, and avoid overfitting to majority classes. 

\subsection{INCLUDE}

INCLUDE \cite{Sridhar2020} is an Indian Sign Language (ISL) dataset that contains 270,000 frames across 4,287 videos over 263 word signs from 15 different word categories. INCLUDE is recorded with the help of experienced signers to provide close resemblance to natural conditions. To maintain data quality, we excluded signs with fewer than 5 samples, resulting in 2,949 sign poses across 253 classes.

This dataset presents different challenges from LSFB, with fewer classes but also fewer samples per class, requiring careful optimization to achieve good performance.

\subsection{DiSPLaY}

The signs were collected at Shahid Beheshti University, Tehran, and show local medical sign language gestures \cite{Ehsan2020}. Microsoft Kinect v2 was used for collecting the multi-modal data, including color frames, depth frames, infrared frames, body index frames, mapped color body on depth scale, and 2D/3D skeleton information. To standardize our body skeleton extraction, we discarded the included skeleton data and used MediaPipe to extract the key points.

This is the smallest and most specialized dataset, requiring a model with good generalization capabilities.

\subsection{Data preprocessing}

For processing, keypoints were formatted in a single dimension as $\{x_i,y_i,z_i|i=0,1,2,...,n\}$ where $n$ represents the number of keypoints in each sample. Additionally, each sample incorporates a temporal dimension, maintaining an equivalent count of keypoints. To facilitate uniform input size, we randomly sample 32 contiguous frames from each clip. For clips comprising fewer than 32 frames, we employ zero padding.

We apply Discrete Cosine Transform (DCT) \cite{Guo2022} preprocessing to enhance temporal feature encoding prior to model input, improving the representation of sequential dynamics.

\section{Methods}

In this section, we outline our methodology. We describe the approach of each component in their corresponding subsection. Section \ref{strategy} introduces our synthetic data generation approach, Section \ref{generative} details our generative model architectures and training, and Section \ref{predictive} describes our predictive models for sign language classification.

\subsection{Pretraining with Synthetic Data} \label{strategy}

Formally, let \(P_r\) denote the underlying discrete distribution of real data samples comprising human body, face, and hand landmarks in the sign language domain. Given a limited labeled dataset \(S_r:=\{(x_1,y_1),...,(x_{n_t},y_{n_t}))\}\sim P_r \) containing \(n_t\) elements, we propose a generative framework to address the inherent data scarcity.

We construct a parametric generative model that approximates a continuous distribution \(P_g\). The training objective minimizes the distributional divergence between the empirical distribution of real training samples \(S_r\) and the learned generative distribution \(P_g\).

To generate class-specific samples, we formulate a conditional generative model \(P_g(X|Y=y)\), where the random variable \(X\) encapsulates the spatiotemporal sign landmark information and \(Y\) represents the corresponding sign class label. This conditional framework facilitates controlled generation of samples belonging to specific sign categories, enabling class-balanced synthetic dataset creation:

\begin{equation}
    S_g\:=\bigcup\limits_{y'=1}^{\mathcal{C}} \{(x'_{1},y'),...,(x'_{n},y'))\}    
\end{equation}

Where \(\mathcal{C}\) is the total number of classes. Each class \(y'\) can have \(n\) distinct generated samples \(x'\), allowing us to create multiple diverse variations of the same sign class. 

We implement a two-phase training protocol for our sign language recognition (SLR) models:

\begin{itemize}
    \item Pretraining Phase: Initially, the model parameters are optimized using the synthetic dataset \(S_g\). This constitutes approximately 10\% of the total training iterations and serves to establish a robust initialization that exhibits enhanced generalization capabilities.
    \item Fine-tuning Phase: Subsequently, we refine the pretrained model using the authentic dataset \(S_r\).
\end{itemize}

\subsection{Generative model} \label{generative}

We develop a conditional human motion generator model (CMLPe) to create new realistic and semantically correct synthetic samples. 
Our model's training employs a two-parts composite loss function similar to that used in SiMLPe \cite{Guo2022}. The total loss minimizes the components: (1) L2-norm between predicted motion and ground-truth motion, and (2) L2-norm between the velocity of the predicted motion and the ground-truth velocity.

\subsubsection{CMLPe}

\begin{figure}[H]
    \centering
    \includegraphics[width=\textwidth,height=\textheight,keepaspectratio]{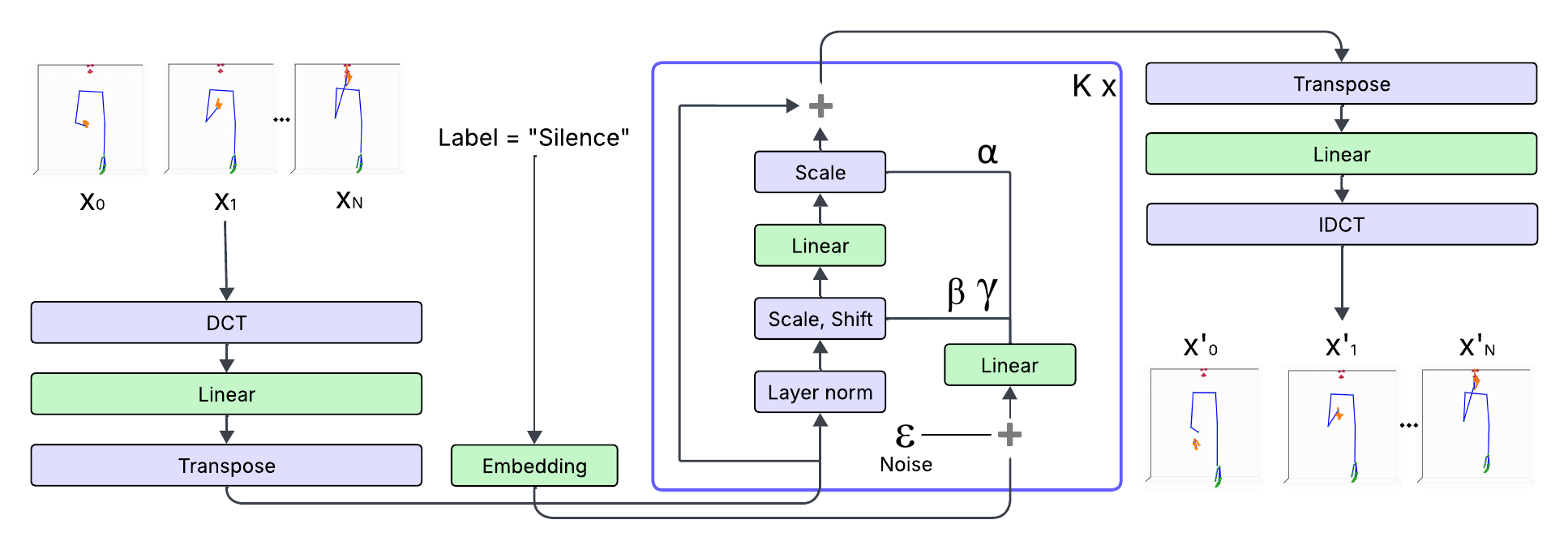}
    \caption{Conditional MultiLayer Perceptron (CMLPe) architecture. The model is mainly composed of linear layers and conditioned with Adaptive Layer Normalization (adaLN) on each of its $K$ blocks.}
    \label{fig:cmlpe}
\end{figure}

Our \textbf{Conditional MultiLayer Perceptron (CMLPe)} based human motion generator, inspired by siMLPe \cite{Guo2022}, is composed of 3 blocks: input (IN), output (OUT) and conditional MLP blocks (CMB). Input and output blocks process spatial features, while CMB works with the temporal dimension. The model architecture is summarized in Figure \ref{fig:cmlpe}, the mathematical notation used in the formal definition is described in Table \ref{tab:model_notation}.

\begin{table}[ht!]
    \centering
    \footnotesize
    \caption{Mathematical notation for the proposed conditional 3D human pose prediction framework. The model takes pose sequences and sign class labels as inputs to predict future pose sequences.}
    \begin{tblr}{
        colspec={l{c}l},
        row{1}={font=\bfseries},
        column{1}={font=\itshape},
        row{even}={bg=gray!10},
    }
        \textbf{Symbol}                  & \textbf{Dimension} & \textbf{Description} \\
        \toprule
        $X$                             & $\mathbb{R}^{N \times L}$ & Full frame sequence of $N$ frames with $L$ pose landmarks \\
        $X^{\text{value}}$              & $\mathbb{R}^{M \times L}$ & Input sequence with $M < N$ frames \\
        $X^{\text{target}}$             & $\mathbb{R}^{T \times L}$ & Target sequence with $T = N - M$ frames \\
        $Y$                             & $\{1, ..., C\}$ & Input sign class label for $C$ classes \\
        $X'$                            & $\mathbb{R}^{T \times L}$ & Predicted future pose sequence \\
        \bottomrule
    \end{tblr}
    \label{tab:model_notation}
\end{table}


First, a Discrete Cosine Transform (DCT) is applied to the input pose in order to encode the temporal information. The input block applies a fully connected layer of size $D$ working on the $L$ spatial features of the input. Then, a transpose operation flips the feature map, allowing the CMB to process temporal information.

\begin{equation}
z^{\text{in}}=\text{IN}(\text{DCT}(X^{\text{value}})) \in \mathbb{R}^{D\times N}    
\end{equation}

We employ multiple CMBs that process temporal information by applying fully connected layers across the encoded frame dimension. CMB is inspired by the Adaptive Layer Normalization (adaLN) blocks used in multiple GAN and diffusion models, in particular to Diffusion Transformers \cite{Peebles2022}. We apply an embedding layer to $Y$ mapping it to $Y’ \in \mathbb{R}^{D}$.  We regress the shift $\gamma$, scale $\beta$ and gate $\alpha$ values from a fully connected layer applied to $Y’$ with added random noise $\epsilon$ to generate variations in the output. Then, for each block, given an input feature map $z’$ we use a layer normalization operation (LN), we apply adaLN using the $\beta$ and $\alpha$. Finally, we proceed with a fully connected layer gated by $\alpha$. Formally:

\begin{equation}
z^{\text{adaLN}}=\text{LN}(z')(1+\beta)+\gamma
\end{equation}

\begin{equation}
z^{\text{CMB}}=\alpha(\text{FC}(z^{\text{adaLN}})) \in \mathbb{R}^{D\times N}
\end{equation}

Finally, in the output block, we transpose the result of the last CMB $z^{CMB’}$ and apply a fully connected layer of size $L$ over the spatial dimension as with the input block. Then we use an IDCT to obtain our prediction:

\begin{equation}
X' = X^{\text{value}}_M \oplus \text{IDCT}(\text{OUT}(z^{\text{CMB}'})) \in \mathbb{R}^{T\times L}
\end{equation}

where $\oplus$ denotes the element-wise addition of each predicted motion frame to the last input frame $X^{\text{value}}_M$.

To generate complete sign sequences $X^{\text{pred}} \in \mathbb{R}^{N \times L}$ we employ a dual-generator approach where $T=N/2$, utilizing two complementary CMLPe models operating on reversed temporal segments. Each of these models generates half of the complete sequence. The input and target of each model differ, one receives $X^{\text{value}}$ as the input and $X^{\text{target}}$ as the objective, while the other reverses the order of the frames and receives $\text{reverse}(X^{\text{target}})$ as the input and $\text{reverse}(X^{\text{value}})$ as the objective. We then join the result of both models to obtain a full generated sequence. In addition, we introduce a mild random Gaussian noise to the label embedding, enabling the generation of multiple variations of the same sign.

\subsection{Predictive models} \label{predictive}

To rigorously evaluate the effectiveness of our synthetic data augmentation approach, we measure its effect on a Sign Language Recognition (SLR) task with two state-of-the-art architectures. First, we employ a Transformer model, which has become the standard for dynamic SLR tasks \cite{Fink2023}. Second, we leverage the recently developed Mamba architecture \cite{mamba}, which has demonstrated exceptional performance through its state space sequence modeling approach while maintaining computational efficiency. Both architectures are trained using a cross-entropy loss function, enabling direct performance comparison under similar training conditions and consistent evaluation of our data augmentation strategy's impact across different modeling paradigms. For both models, the input pose data is tokenized at the spatial dimension by mapping each spatial vector into a discrete token representation, while preserving the original temporal structure. To this end, both models implement linear layers for embedding.

\begin{figure}[ht!]
    \centering
    \begin{subfigure}{0.45\textwidth}
        \includegraphics[width=\hsize]{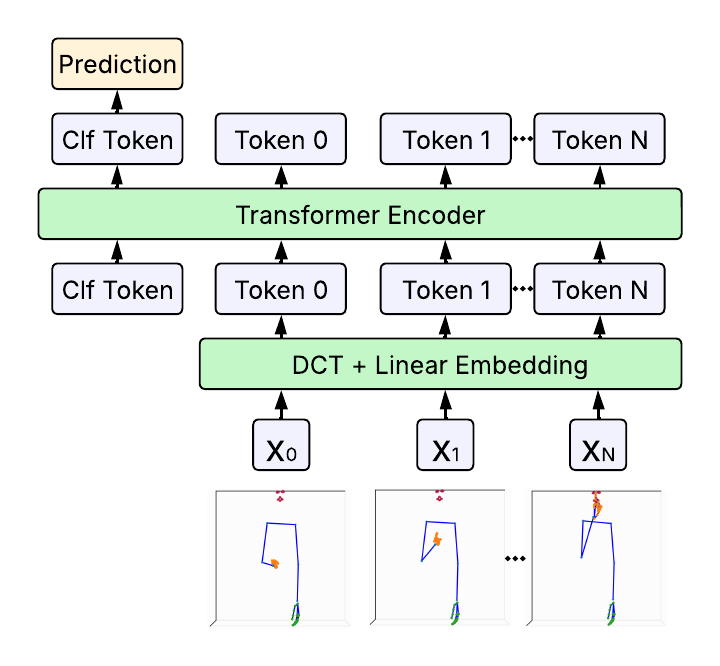}
        \caption{Transformer-SL}
        \label{fig:transformer-sl}
    \end{subfigure}
    \hfill
    \begin{subfigure}{0.45\textwidth}
        \includegraphics[width=\hsize]{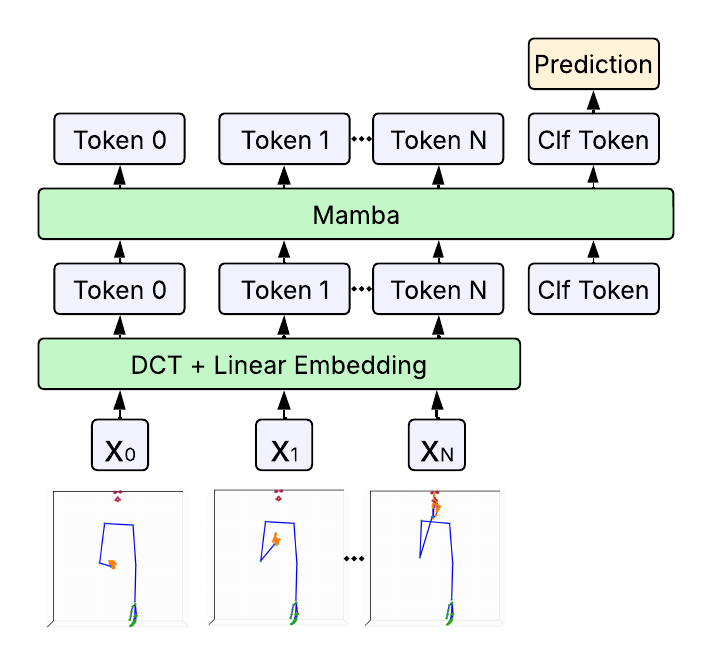}
        \caption{Mamba-SL}
        \label{fig:mamba-sl}
    \end{subfigure}
    \caption{
    Transformer-SL and Mamba-SL architecture for SLR. A classification token is concatenated to the sequence. Then is passed to the backbone along the input and the resulting classification token is used to predict the label. Note that this token is located at the end of the sequence in Mamba-SL following its sequential process.}
    \label{fig:slr_models}
\end{figure}

\subsubsection{Transformer-SL}
Our Transformer-SL model is inspired by the Vision Transformer (ViT) \cite{Dosovitskiy2021} architecture for classification, in which a classification token is concatenated at the beginning of the sequence. The features are then encoded by using a linear layer and adding positional encoding to each token. Both the classification token and positional encoding are 1D trainable vectors. These tokens are then processed by multiple residual transformer encoder blocks. Finally, the classification token is passed to a multilayer perceptron containing a normalization layer followed by a linear layer in order to output the model's prediction. Figure \ref{fig:transformer-sl} summarizes this process.

\subsubsection{Mamba-SL}
For our Mamba-SL model, we made a series of modifications on the original Mamba architecture \cite{mamba} to adapt the model for SLR tasks. We introduce a linear layer on the input instead of an embedding layer to encode the input tokens. We introduce a trainable parameter to the end of the token chain as a class token that is later used for classification. We use the same backbone as the original Mamba model, with simplified state space model (SSM) blocks. Then we use the class token to make our prediction using a normalization and a linear layer.
Note that during training and inference, the final token of each sample’s frame sequence may vary. To prevent the padding from altering the prediction of our model, we add the class token to the end of the sequence, after the last token that comes before the padding. The final linear layer of the model takes as input this classifier token using a mask to locate it efficiently. In contrast with our Transformer-SL model, the token must be placed at the end of the sequence due to the sequential nature of Mamba-SL, where each token passes information to the next in the sequence. Figure \ref{fig:mamba-sl} summarizes this process.

\section{Experiments and results}

In this section, we present the procedure, results, and analysis for each experiment. We describe the metrics used in each corresponding subsection. Section \ref{experiment-generation} displays the results of our sign language generation experiments using CMLPe, Section \ref{experiment-recognition} shows our experiments using our SLR models with and without synthetic data augmentation, and Section \ref{experiment-da} is our comparison of the usage of synthetic data augmentation against other data augmentation methods.

\subsection{Sign language generation}
\label{experiment-generation}

For our generation experiments, we employ our CMLPe model with a consistent architecture across all three datasets: 6 MLP blocks with an embedding size of 32 and a noise scale of 0.1 to enhance output variability. Each sample consists of 32 frames (16 input, 16 target output). Training uses an Adam optimizer with a learning rate of $1\times10^{-4}$, weight decay of $1\times10^{-4}$, and 100 training steps. To generate full sign sequences, we train a reversed generator, reversing input-target relationships in order to create the first half of the synthetic samples (the half that served as input to the regular CMLPe model). To address class imbalance, we oversample by replicating samples from underrepresented classes to match the modal frequency.

Our primary evaluation metric is Mean Per Joint Position Error (MPJPE), which quantifies the average L2 distance between predicted and ground-truth joint positions. Table \ref{tab:MPJPE} displays the performance for each dataset. To our knowledge, we are the first to present SLG of pose sequences on these datasets. In particular, our model achieves its best performance in the LSFB dataset, with similar performance in the INCLUDE and DiSPLaY datasets. LSFB's superior results likely stem from its more diverse training dataset, which enables the model to better learn the varied signing patterns present within it. The table also shows a significant difference between the MPJPE of the regular and reversed versions of the generator model. This disparity is caused by the varied length of the data input and the right padding used, as there are multiple instances of signs with fewer than 32 frames in LSFB.

Qualitative assessment through visual inspection confirms that our generated sequences preserve semantic meaning while introducing naturalistic variations in movement dynamics. This variation is necessary for effectively augmenting the SLR training dataset.

\begin{table}[ht!]
    \centering
    \footnotesize
    \caption{Mean Per Joint Position Error (MPJPE) for our sign generation models across different datasets. Lower values indicate better performance with more accurate joint position predictions.}
    \begin{tblr}{
        colspec={lll},
        row{1}={font=\bfseries},
        column{1}={font=\itshape},
        row{even}={bg=gray!10},
    }
        \textbf{Dataset}                   & \textbf{CMLPe} & \textbf{reversed-CMLPe} \\
        \toprule
        LSFB                               & 14.1                 & \textbf{8.8} \\
        INCLUDE                            & \textbf{25.7}                 & 26.5 \\
        DiSPLaY                            & \textbf{24.6}                 & 26.9 \\
        \bottomrule
    \end{tblr}
    \label{tab:MPJPE}
\end{table}

\subsection{Sign language recognition}
\label{experiment-recognition}

To evaluate the effectiveness of our synthetic data augmentation approach, we conduct experiments with Transformer-SL and Mamba-SL across three datasets. Several preprocessing techniques are employed in order to optimize model performance:

\begin{itemize}
    \item Padding method: We compare zero padding with other padding strategies, finding that zero padding performs best in Transformer-SL when combined with attention masking to prevent the model from focusing on padded frames. In Mamba-SL, the padding strategy does not have a considerable effect on the model performance due to the causal nature of the class token.
    \item Window size: We experiment with different temporal window sized to capture the full motion of signs while minimizing unnecessary frames. Although sign sequence size varies between datasets, a window size of 32 frames provides the best balance for our 3 datasets, resulting in the best performance.
    \item Oversampling: To address class imbalance, we implement class-balanced oversampling during training, ensuring that each sign class contributed equally to the loss function regardless of its frequency in the dataset by replicating samples from underrepresented classes to match the modal frequency.
\end{itemize}

All models employ the RAdam optimizer \cite{Liu2019} coupled with the OneCycle learning rate scheduling strategy \cite{smith2019super} to ensure stable and efficient convergence. This learning rate strategy allows our models to train in a lower number of steps without overfitting. For regularization, we apply a dropout rate of 0.2 and weight decay. Transformer-SL and Mamba-SL configuration details for each dataset are displayed in Tables \ref{tab:transformer_config} and \ref{tab:mamba_config}.

\begin{table}[ht!]
    \centering
    \footnotesize
    \caption{Transformer-SL model configuration across datasets.}
    \begin{tblr}{
        colspec={lccc},
        row{1}={font=\bfseries},
        column{1}={font=\itshape},
        row{even}={bg=gray!10},
    }
        Parameter                 & LSFB     & INCLUDE & DiSPLaY \\
        \toprule
        Layers                   & 2        & 2       & 2 \\
        Attention Heads          & 4        & 4       & 4 \\
        Hidden Dimension         & 80       & 80      & 80 \\
        MLP Dimension            & 256      & 128     & 128 \\
        Output Size              & 1024     & 1024    & 1024 \\
        Batch Size               & 2048     & 16      & 16 \\
        Weight Decay            & $1\times10^{-4}$ & $1\times10^{-4}$ & $1\times10^{-4}$ \\
        Learning Rate (peak)     & $1\times10^{-3}$ & $1\times10^{-2}$ & $1\times10^{-2}$ \\
        Training Steps           & 50       & 400     & 400 \\
        Warmup Ratio             & 30\%     & 30\%    & 10\% \\
        Synthetic Pretraining Steps & 5     & 75      & 50 \\
    \end{tblr}
    \label{tab:transformer_config}
\end{table}

\begin{table}[ht!]
    \centering
    \footnotesize
    \caption{Mamba-SL model configuration across datasets.}
    \begin{tblr}{
        colspec={lccc},
        row{1}={font=\bfseries},
        column{1}={font=\itshape},
        row{even}={bg=gray!10},
    }
        Parameter                 & LSFB     & INCLUDE & DiSPLaY \\
        \toprule
        Layers                   & 1        & 2       & 1 \\
        Hidden Dimension         & 512      & 64      & 64 \\
        Output Size              & 1024     & 1024    & 1024 \\
        Batch Size               & 2048     & 16      & 16 \\
        Weight Decay            & $1\times10^{-3}$ & $1\times10^{-4}$ & $1\times10^{-4}$ \\
        Learning Rate (peak)     & $1\times10^{-4}$ & $1\times10^{-2}$ & $1\times10^{-2}$ \\
        Training Steps           & 50       & 400     & 400 \\
        Warmup Ratio             & 10\%     & 30\%    & 10\% \\
        Synthetic Pretraining Steps & 5     & 75      & 50 \\
    \end{tblr}
    \label{tab:mamba_config}
\end{table}

In the following subsections, we analyze the results of the model in each of the three datasets.

\subsubsection{LSFB}

Table \ref{tab:lsfb_sota_comparison} demonstrates that both Transformer-SL and Mamba-SL architectures achieve substantial performance gains through synthetic data pretraining on the LSFB dataset. It improves the performance by 1.3\% for Mamba-SL (53.2\% to 54.5\%) and 1.2\% for Transformer-SL (53.8\% to 55.0\%) when compared with the same model architectures trained using only real data.

Our synthetic data pretraining approach establishes new state-of-the-art performance on the LSFB dataset, surpassing previous methods \cite{Fink2023,Rios2024}. Our Transformer-SL model with synthetic pretraining achieves an accuracy of 55.0\%, surpassing the previous state-of-the-art Transformer-based model with 54.4\% accuracy by 0.6\%. This improvement demonstrates that our generated data effectively captures the nuances of French Belgian Sign Language, helping the model generalize despite limited training data.

\begin{table}[ht!]
    \centering
    \footnotesize
    \caption{Accuracy comparison of our models trained with real data only and pretraining with synthetic data against state-of-the-art approaches on the LSFB dataset. Bold values indicate the best overall performance.}
    \begin{tblr}{
        colspec={l{c}l},
        row{1}={font=\bfseries},
        column{1}={font=\itshape},
        row{even}={bg=gray!10},
        hline{6} = {1}{-}{},
        hline{6} = {2}{-}{},
     }
        \textbf{Model} & \textbf{Synthetic pretraining} & \textbf{Accuracy \%} \\
        \toprule
        Mamba-SL (Ours) & \xmark & 53.2 \\
        Mamba-SL (Ours) & \cmark & \textbf{54.5} \\
        Transformer-SL (Ours) & \xmark & 53.8 \\
        Transformer-SL (Ours) & \cmark & \textbf{55.0} \\
        Transformer \cite{Fink2023} & \xmark & 54.4 \\
        ConvAtt \cite{Rios2024} & \xmark & 42.7 \\
        \bottomrule
    \end{tblr}
    \label{tab:lsfb_sota_comparison}
\end{table}

\subsubsection{INCLUDE}

As presented in Table \ref{tab:include_sota_comparison}, both architectures show consistent improvement by using synthetic data on the INCLUDE dataset. The Mamba-SL architecture exhibits a more substantial gain of 3.1\%, improving from 81.9\% to 85.0\%. The Transformer-SL model achieves the highest overall accuracy of 87.1\%, representing a 0.7\% improvement over training with real data only.

When comparing with existing approaches in Table \ref{tab:include_sota_comparison}, our Transformer-SL model's 87.1\% accuracy is competitive with LSTM-based approaches \cite{Khartheesvar2024,Sridhar2020} (87.4\% and 85.6\%) but falls short of specialized graph-based architectures like SL-GCN \cite{Selvaraj2022} (93.5\%) and HWGAT \cite{Patra2024} (97.7\%). This suggests that while exploiting our synthetic data is effective, the architectural advantages of graph-based models for capturing spatial relationships between body parts remain significant for this dataset.

\begin{table}[ht!]
    \centering
    \footnotesize
    \caption{Accuracy comparison of our models trained with real data only and pretraining with synthetic data against state-of-the-art approaches on the INCLUDE dataset. Bold values indicate the best overall performance.}
    \begin{tblr}{
        colspec={l{c}l},
        row{1}={font=\bfseries},
        column{1}={font=\itshape},
        row{even}={bg=gray!10},
        hline{6} = {1}{-}{},
        hline{6} = {2}{-}{},
    }
        \textbf{Model} & \textbf{Synthetic pretraining} & \textbf{Accuracy \%} \\
        \toprule
        Mamba-SL (Ours)                  & \xmark & 81.9 \\
        Mamba-SL (Ours)                  & \cmark & 85.0 \\
        Transformer-SL (Ours)            & \xmark & 86.4 \\
        Transformer-SL (Ours)            & \cmark & 87.1 \\
        LSTM \cite{Khartheesvar2024}    & \xmark & 87.4 \\
        LSTM \cite{Sridhar2020}         & \xmark & 85.6 \\
        SL-GCN \cite{Selvaraj2022}      & \xmark & 93.5 \\
        HWGAT \cite{Patra2024}          & \xmark & \textbf{97.7} \\
        \bottomrule
    \end{tblr}
    \label{tab:include_sota_comparison}
\end{table}

\subsubsection{DiSPLaY}

The DiSPLaY dataset shows the most dramatic improvements from using synthetic data. As shown in Table \ref{tab:DiSPLaY_real_vs_synthetic}, our Mamba-SL model achieves a remarkable 9.0\% accuracy increase (from 88.3\% to 97.3\%) when pretrained with synthetic data. The Transformer-SL model also shows substantial improvement, increasing from 92.8\% to 95.5\% (an increase of 2.7\%).

No previous SLR results were found for this dataset. Therefore, our Mamba-SL model with synthetic data augmentation is positioned as the state-of-the-art for the DiSPLaY dataset with 97.3\% accuracy. The exceptional performance gains on this dataset when using synthetic data result from several contributing factors:

\begin{itemize}
    \item The medical signs in DiSPLaY have more standardized movements with less dialectal variation, making them easier to generate synthetically with high fidelity.
    \item The smaller size and class count of the dataset make it particularly susceptible to benefits from enhancing its data distribution.
    \item Mamba-SL's sequential processing appears to be especially well-suited for capturing the temporal dynamics of medical signs.
\end{itemize}

The substantial improvement observed in both architectures underscores the effectiveness of our synthetic data pretraining approach, particularly for specialized domains with limited training data.

\begin{table}[ht!]
    \centering
    \footnotesize
    \caption{Accuracy comparison of our models trained with real data only and pretraining with synthetic data on the DiSPLaY dataset. Bold values indicate the best overall performance.}
    \begin{tblr}{
        colspec={l{c}l},
        row{1}={font=\bfseries},
        column{1}={font=\itshape},
        row{even}={bg=gray!10},
    }
        \textbf{Model}                  & \textbf{Synthetic pretraining} & \textbf{Accuracy \%} \\
        \toprule
        Mamba-SL                         & \xmark & 88.3 \\
        Mamba-SL                         & \cmark & \textbf{97.3} \\
        Transformer-SL                   & \xmark & 92.8 \\
        Transformer-SL                   & \cmark & \textbf{95.5} \\
        \bottomrule
    \end{tblr}
    \label{tab:DiSPLaY_real_vs_synthetic}
\end{table}

\subsubsection{Data augmentation vs Synthetic Data Generation}
\label{experiment-da}

We evaluated traditional data augmentation techniques both with and without synthetic data to explore their impact and interaction. For these experiments, our data augmentation pipeline includes random rotations ($\pm5^\circ$) and scaling variations ($\pm5\%$), applied during training to introduce variability into the real samples. All our experiments with data augmentation are made with the LSFB dataset as it proves to be the hardest task.

As shown in Table \ref{tab:data_augmentation_impact}, data augmentation methods provide modest improvements for most configurations. The Transformer-SL model shows consistent gains from both data augmentation and synthetic pretraining, with accuracy improving from 53.8\% to 54.3\% with data augmentation alone, and from 54.1\% to 55.0\% when combining data augmentation with synthetic data pretraining. This suggests a complementary relationship between data augmentation and synthetic pretraining for the Transformer-SL architecture.

The Mamba-SL model displays a different behavior. While data augmentation alone doesn't result in a significant improvement (52.9\% to 53.0\%), the model with synthetic data performs worse when traditional augmentation is added (54.5\% to 54.1\%). This suggests that the diversity introduced by synthetic samples already provides sufficient regularization.

\begin{table}[ht!]
    \centering
    \footnotesize
    \caption{Classification accuracy on the LSFB dataset comparing models trained with real data only and using synthetic data. We evaluate each approach with and without data augmentation (DA). Bold values indicate the best performance for each model configuration.}
    \begin{tblr}{
        colspec={l{c}ll},
        row{1}={font=\bfseries},
        column{1}={font=\itshape},
        row{even}={bg=gray!10},
    }
        \textbf{Model}                  & \textbf{Synthetic pretraining}& \textbf{No DA}    & \textbf{DA} \\
        \toprule
        Transformer-SL                     & \xmark & 53.8              & \textbf{54.3} \\
        Mamba-SL                           & \xmark & 52.9              & \textbf{53.0} \\
        Transformer-SL                     & \cmark & 54.1              & \textbf{55.0} \\
        Mamba-SL                           & \cmark & \textbf{54.5}     & 54.1 \\
        \bottomrule
    \end{tblr}
    \label{tab:data_augmentation_impact}
\end{table}

\section{Conclusions \& Future Work}

The proposed CMLPe generator leverages a computationally efficient architecture to synthesize high-fidelity sign language samples across three diverse datasets (LSFB, INCLUDE and DiSPLaY) with minimal computational resources, proving the practical applicability of generating new synthetic datasets, even for research teams with limited computing infrastructure.

We developed two Sign Language Recognition (SLR) architectures, Mamba-SL and Transformer-SL, which are able to extract temporal and local information of pose data to correctly predict single sign sequences. Using our conditional MLP-based network for human motion prediction (CMLPe), we were able to consistently improve recognition accuracy for SLR by pretraining on generated samples. Our experiments establish a new state-of-the-art in SLR for LSFB and DiSPLaY, improving over the previous state-of-the-art by 0.6\% for LSFB and establishing a baseline for DiSPLaY which lacked previous works in this field.

Although both recognition architectures demonstrate consistent gains from synthetic data pretraining, Mamba-SL exhibits superior performance improvements across all evaluated datasets, in particular on the DiSPLaY dataset where this technique improved the accuracy by 9.0\%. This could be attributed to the difference in the number of parameters, making synthetic data a key component in preventing the model from overfitting.

In addition, synthetic data generation showed compatibility with traditional augmentation approaches, leveraging the benefits of both techniques. 

For future work, the SLG model could be further improved by leveraging data from multiple sign language or gesture datasets for unsupervised learning. This method could reduce the need for in-distribution data for each dataset. In addition, a sequential approach could be taken for the generation, exchanging computational cost for better quality samples; this would replace the dual generation of the CMLPe and the reversed-CMLPe generators.

\section*{CRediT authorship contribution statement}
\textbf{Gaston Gustavo Rios:} Conceptualization, Data curation, Investigation, Methodology, Software, Validation, Visualization, Writing – original draft

\textbf{Pedro Dal Bianco:} Conceptualization, Methodology

\textbf{Franco Ronchetti:} Conceptualization, Formal Analysis, Methodology, Supervision, Writing – review \& editing

\textbf{Facundo Quiroga:} Conceptualization, Formal Analysis, Methodology, Supervision, Writing – review \& editing

\textbf{Oscar Stanchi:} Methodology, Writing – review \& editing

\textbf{Santiago Ponte Ahón:} Methodology, Writing – review \& editing

\textbf{Waldo Hasperué:} Project administration, Resources

\section*{Declaration of competing interest}
The authors declare that they have no known competing financial interests or personal relationships that could have appeared to influence the work reported in this paper.

\section*{Data availability}
All our research data comes from public datasets. Synthetic datasets can be generated by using the specified models.








\bibliographystyle{elsarticle-harv} 
\bibliography{bibliography.bib}





\end{document}